\documentclass[letterpaper, 10 pt, conference]{ieeeconf}  

\IEEEoverridecommandlockouts                              
    
\usepackage{algorithm}
\usepackage{algpseudocode}
\usepackage{setspace}
\usepackage{hyperref}
\usepackage{graphics} 
\usepackage{graphicx}
\usepackage{pgfplots}
\pgfplotsset{
    width=1\columnwidth,
    compat=1.9,
    every axis/.append style={
        label style={font=\small},
        tick label style={font=\small}  
        },
    every semilogyaxis/.append style={
        label style={font=\small},
        tick label style={font=\small}  
        }
}

\usepgfplotslibrary{external}
\usepackage{xspace}
\makeatletter
\DeclareRobustCommand\onedot{\futurelet\@let@token\@onedot}
\def\@onedot{\ifx\@let@token.\else.\null\fi\xspace}

\usepackage[final]{changes}
\definechangesauthor[name={TODO}, color=red]{TODO}
\definechangesauthor[name={Michal}, color=blue]{MV}
\definechangesauthor[name={Rado}, color=blue]{RS}
\definechangesauthor[name={Gabina}, color=blue]{GS}
\definechangesauthor[name={Karla}, color=blue]{KS}

\newcommand{\rddl}[0]{{PRAG}\xspace} 

\title{\LARGE \bf
PRAG: Procedural Action Generator
}

\author{{Radoslav \v{S}koviera$^1$, Michal Vavre\v{c}ka$^1$, Gabriela \v{S}ejnov\'{a}$^1$, Karla \v{S}t\v{e}p\'{a}nov\'{a}$^1$ }
\thanks{$^{1}$CIIRC, Czech Technical University in Prague
        {\tt\small michal.vavrecka, radoslav.skoviera, gabriela.sejnova, karla.stepanova@cvut.cz}}%
\thanks{M.V. and R.\v{S}. contributed equally to the paper and were supported by the Czech Science Foundation (GACR) grant no. 23-04080L, G.\v{S}. and K.\v{S}. by GACR grant no.21-31000S. R.\v{S}. and K.\v{S}. were further supported by the European Union under the project ROBOPROX (reg. no. CZ.02.01.01/00/22\_008/0004590).}
}

\begin{document}

\maketitle
\thispagestyle{empty}
\pagestyle{empty}



\begin{abstract}
We present a novel approach for the procedural construction of multi-step contact-rich manipulation tasks in robotics. Our generator takes as input user-defined sets of atomic actions, objects, and spatial predicates and outputs solvable tasks of a given length for the selected robotic environment. The generator produces solvable tasks by constraining all possible (nonsolvable) combinations by symbolic and physical validation. The symbolic validation checks each generated sequence for logical and operational consistency, and also the suitability of object-predicate relations. Physical validation checks whether tasks can be solved in the selected robotic environment. Only the tasks that passed both validators are retained. The output from the generator can be directly interfaced with any existing framework for training robotic manipulation tasks, or it can be stored as a dataset of curated robotic tasks with detailed information about each task. This is beneficial for RL training as there are dense reward functions and initial and goal states paired with each subgoal. It allows the user to measure the semantic similarity of all generated tasks. We tested our generator on sequences of up to 15 actions resulting in millions of unique solvable multi-step tasks.
\end{abstract}

\section{INTRODUCTION}

Human and robotic learning are dependent on the ability to generate and solve novel tasks, progressively building more complex skills over time. However, in robotics, the scarcity of diverse and sufficiently rich training data presents a major challenge for reinforcement learning (RL). Traditionally, RL tasks are designed manually \cite{mahadevan, breyer, rlbench}, a process that is labor intensive and inherently constrained in complexity and scalability. This limited task repertoire restricts the ability of a robot to generalize to new and uncounted scenarios.

\begin{figure}[t]
    \centering
    \includegraphics[width=0.9\columnwidth]{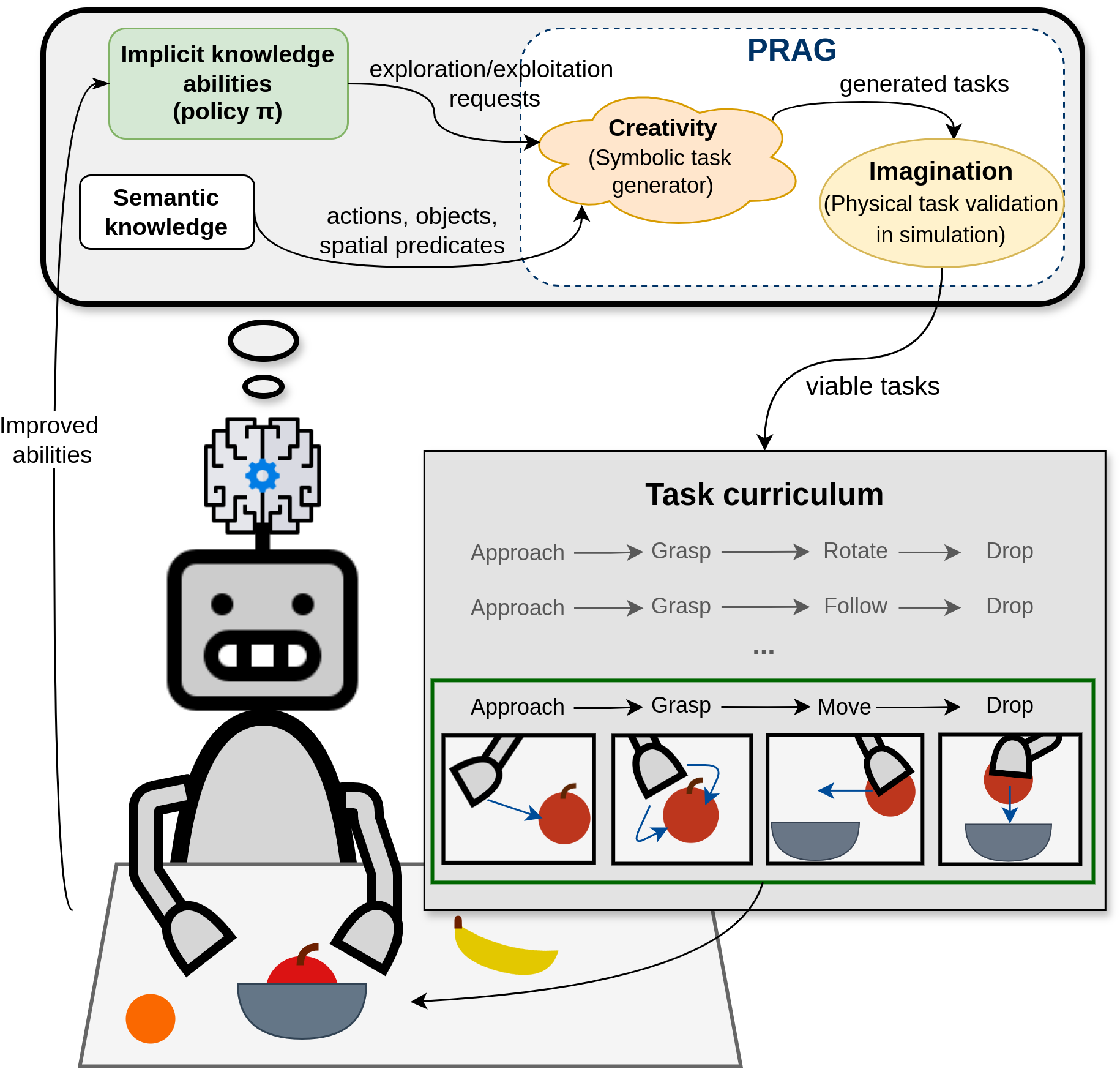} 
    \caption{General scheme of the proposed PRAG system. PRAG takes semantic knowledge (actions, objects, predicates) and generates a curriculum of viable tasks through a two-stage creativity (symbolic) and imagination (physical) process, which in turn improves the robot's abilities.}
    \label{fig1}
\end{figure}

To tackle this challenge, we developed a novel Procedural Action Generator (\rddl), capable of creating long-horizon contact-rich manipulation tasks of arbitrary length (see Fig.~\ref{fig1}). It extends recent methods based on intrinsic motivation with mechanisms for novel task generation based on principles of combinatorial composition. The main advantages of our approach are: 
\begin{itemize}
    \item Ability to generate a large set of tasks from a small set of pre-defined constraints. The generated tasks follow the PDDL~\cite{fox2006modelling} nomenclature, allowing the use of standard planners.
    \item Use of abstract classes for skills and objects, allowing for easy specification and integration with robotic simulators.
    \item Adaptive control over the balance between exploration (generating diverse tasks) and exploitation (focusing on specific skills or objects).
\end{itemize}

Related works in robotics procedural task generation, such as POET \cite{poet} or ATR \cite{fang3}, often focus on generating new environments or randomizing parameters for a given set of skills. In contrast, \rddl procedurally constructs the task sequence itself, ensuring that each generated task has at least one symbolically valid solution path. This allows our approach to vary not only skill parameters but also the sequence of actions, enabling agents to generalize across a broader range of long-horizon tasks.

\section{METHODOLOGY}

We define a manipulation task as a sequence of atomic actions (AAs), such as \textit{Approach}, \textit{Grasp}, or \textit{Move}. Our goal is to procedurally generate diverse and feasible sequences of these AAs to create a rich training curriculum. The \rddl system achieves this through a two-stage validation process: symbolic generation and physical validation.

\begin{figure*}[t]
    \centering
    \includegraphics[width=\textwidth]{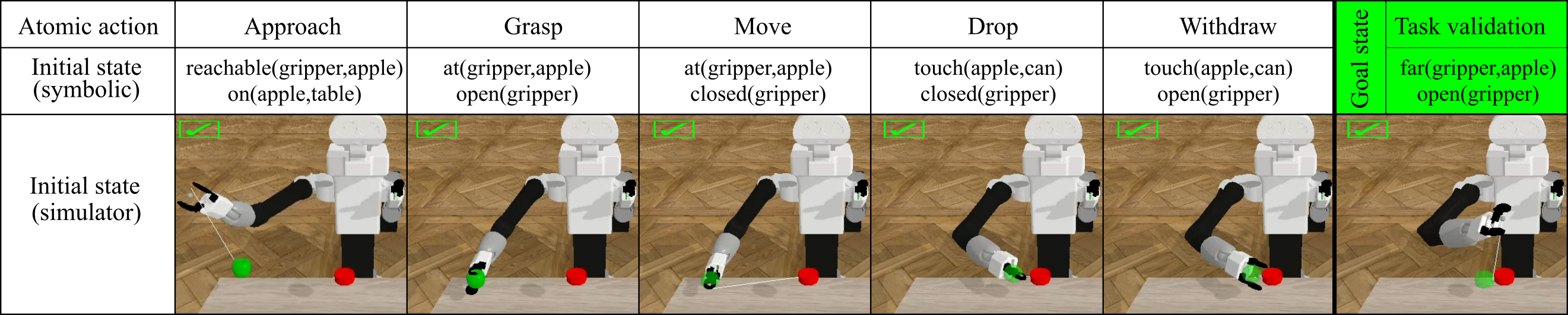}
    \caption{Example of a generated sequence resembling a Pick and Place task. The top row shows the symbolic validation, where each step's initial state (predicates) must be valid. The bottom row shows the physical validation in a simulator, where each state is spawned and checked for physical feasibility. The final goal state is shown in the last column.}
    \label{sim_sequence}
\end{figure*}

\subsection{Symbolic Generation and Validation}

The symbolic generator takes a set of user-defined inputs:
\begin{itemize}
    \item \textbf{Atomic Actions (AAs):} Each AA is defined by preconditions ($C_{init}$) and postconditions ($C_{post}$), which are logical formulas over predicates. For an action to be executable, its $C_{init}$ must be true. After execution, the world state is updated according to $C_{post}$.
    \item \textbf{Entities and Predicates:} Objects (e.g., \textit{apple}, \textit{table}) are organized in a class hierarchy (e.g., \textit{apple} is a \textit{GraspableObject}). The predicates describe the relations between them (e.g., \textit{OnTop(a,b)}, \textit{Holding(g,o)}).
    \item \textbf{Sampling Weights:} Weights can be applied to actions, action pairs, and entities to bias the generation process toward the exploration or exploitation of specific skills.
\end{itemize}

The generation algorithm (inspired by the PDDL and SAT solvers) works by iteratively building a sequence of AAs. For each step, it samples an action, instantiates required objects (either by linking existing objects or creating new ones from the hierarchy), and validates its preconditions ($C_{init}$). If valid, the action is added to the sequence, its postconditions ($C_{post}$) are applied to the symbolic world state, and the process continues. This ensures that every generated sequence is logically and operationally consistent. This method drastically prunes the search space; for a sequence of 15 actions with 8 AAs, there are more than $3.5 \times 10^{13}$ total combinations, but only about $1.9 \times 10^8$ are symbolically valid.

\subsection{Physical Validation in Simulation}

A symbolically valid task may still be physically impossible (e.g. due to robot reach limitations or object collisions). Therefore, each generated task is passed to a physical simulator (we use a modified myGym simulator \cite{mygym}) for validation. For each AA in the sequence, the simulator attempts to spawn the initial and goal states described by the predicates. We use a volume-based spawning method in which spatial predicates (e.g. \textit{left of}) define valid placement volumes for objects. The simulator's physics engine checks for collisions and reachability. Only tasks where every subgoal state in the sequence is physically achievable are considered "viable" and added to the final training curriculum (see Fig.\ref{sim_sequence}).

\section{RESULTS}

We evaluated \rddl's ability to generate valid and diverse tasks. We configured the system with 8 atomic actions and generated sequences of varying lengths.

\subsection{Object Spawning and Physical Feasibility}
We compared our volume-based spawning in myGym against the text-to-image model DALL-E 2 \cite{dalle2} to generate scenes from predicate descriptions. Each model was prompted to create scenes with specific object arrangements. Our method, which directly interprets predicates as geometric constraints, far outperforms DALL-E 2, especially when multiple objects and spatial relations are involved. This highlights the advantage of a structured, geometry-aware approach over a general-purpose generative model for this task. Overall, in a test of 10,000 generated sequences of 3-6 actions, 78.3\% passed our full physical validation pipeline.

\subsection{Advantages for RL Training}
The tasks generated by \rddl are suitable for training RL agents, particularly for long-range problems. The key advantages are:
\begin{itemize}
    \item \textbf{Structured Curriculum:} The generator provides a sequence of subgoals, each with defined initial/goal states and rewards. This enables curriculum learning with dense rewards initially, which can be faded to sparse rewards as the agent's proficiency increases.
    \item \textbf{Guaranteed Solvability:} Every task has at least one known (symbolically valid) solution path. This prevents the agent from wasting resources on impossible tasks, a common issue with unrestricted random scene generation.
    \item \textbf{Efficiency:} By performing symbolic validation first, \rddl avoids the computationally expensive process of running a planner or simulator on exponentially many infeasible task variations.
\end{itemize}

\section{CONCLUSION}
We introduced PRAG, a novel procedural generator for creating diverse, solvable, and long-horizon manipulation tasks. By combining symbolic validation for logical consistency with physical validation for feasibility, PRAG efficiently generates a rich curriculum of tasks from a small set of base components. This approach not only provides structured data for effective RL training, but also serves as a model for how autonomous agents might create their own learning challenges, mirroring principles of human cognitive development. The open source code is available at \href{https://github.com/incognite-lab/prag}{https://github.com/incognite-lab/prag}.

\addtolength{\textheight}{-3cm}   

\end{document}